\pdfoutput=1
\documentclass{article}
\usepackage[final]{colm2025_conference}
\renewcommand{\lhead}[1]{}

\usepackage{microtype}
\usepackage{helvet}%
\usepackage{hyperref}
\usepackage{url}
\usepackage[T1]{fontenc}
\usepackage{graphicx}
\usepackage{subcaption}
\usepackage[frozencache]{minted}
\usepackage[most]{tcolorbox}
\usepackage{xspace}
\usepackage{xstring}
\usepackage{fvextra}
\usepackage{tikz}
\usetikzlibrary{tikzmark,calc,positioning,shadows,shapes,arrows.meta,fit,patterns,backgrounds}
\usepackage{booktabs}
\usepackage{colortbl}
\definecolor{tabgroup}{gray}{0.90}
\usepackage{multirow}
\usepackage{tabularx}
\usepackage{siunitx}
\usepackage{adjustbox}
\usepackage{array}
\usepackage{pifont}%
\usepackage{enumitem}%
\usepackage{amsmath}

\definecolor{darkblue}{rgb}{0, 0, 0.5}
\hypersetup{colorlinks=true, citecolor=darkblue, linkcolor=darkblue, urlcolor=darkblue}

\usepackage{amsmath,amsfonts,bm}

\def\eqref#1{equation~\ref{#1}}

\def\1{\bm{1}}

\DeclareMathAlphabet{\mathsfit}{\encodingdefault}{\sfdefault}{m}{sl}
\SetMathAlphabet{\mathsfit}{bold}{\encodingdefault}{\sfdefault}{bx}{n}

\newcommand{\Lite}{CUA-Lite}
\newcommand{\lgym}{Lite.Gym}
\newcommand{\lsample}{Lite.Sample}
\newcommand{\hpart}[1]{\code{#1}}
\newcommand{\losworld}{Lite.OSWorld}
\newcommand{\lscalecua}{Lite.ScaleCUA}
\newcommand{\lcuagym}{Lite.CUAGym}
\newcommand{\lcuaworld}{Lite.CUAWorld}

\newtcolorbox{codebox}{enhanced, colback=black!2, colframe=black!18, boxrule=0.3pt, arc=2pt, left=2pt, right=2pt, top=0pt, bottom=0pt, boxsep=0pt}

\definecolor{vscbg}{HTML}{FFFFFF}%

\newtcolorbox{takeaway}[1][Takeaway]{%
  colback=black!4!white,
  colframe=black!60!white,
  fonttitle=\bfseries\small,
  colbacktitle=black!80!white,
  coltitle=white,
  enhanced,
  attach boxed title to top left={yshift=-2.2mm, xshift=4mm},
  boxrule=0.6pt,
  arc=1.5pt,
  left=6pt, right=6pt, top=6pt, bottom=4pt,
  title={#1},
}

\tcbset{
  vscodebox/.style={
    width=0.8\linewidth,%
    colback=vscbg,%
    boxrule=1pt,%
    arc=0pt,%
    left=10pt, right=10pt,%
    top=3pt, bottom=0pt,%
    drop shadow={black!50!white},%
    enhanced,%
  }
}

\newenvironment{vscodebox}[1][]{
  \begin{tcolorbox}[vscodebox,#1]
}{
  \end{tcolorbox}
}

\newminted[pycode]{python}{
  fontsize=\small,%
  autogobble,%
  breaklines,%
  python3,%
  frame=none,%
  tabsize=4,%
  stripnl=false,%
  escapeinside=||,%
  texcomments=true,%
}

\newminted[jsoncode]{json}{
  fontsize=\footnotesize,
  autogobble,
  breaklines,
  frame=none,
  tabsize=2,
  stripnl=false,
  escapeinside=||,%
}

\definecolor{diverge}{HTML}{B0563C}%
\definecolor{diffadd}{HTML}{22863A}%
\definecolor{diffdel}{HTML}{CF222E}%
\definecolor{diffaddbg}{HTML}{E6FFEC}%
\definecolor{diffdelbg}{HTML}{FFECEC}%

\newif\ifcameraready%

\newcommand{\code}[1]{\inlinecode{#1}}%

\title{\raisebox{-0.12em}{\includegraphics[height=0.95em]{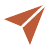}}\,CUA-Lite: An Open Platform for Computer Use Agents}

\author{\hspace*{-\tabcolsep}\rlap{\parbox[t]{\textwidth}{\centering%
Zhanhui~Zhou$^{\dagger}$, Weichen~Zhang, Haoran~Liu, Lingjie~Chen \\
Tianneng~Shi, Kevin~Lin, Zhengyuan~Yang, Lijuan~Wang, Dawn~Song \\[6pt]
\normalfont\href{https://cua-lite.github.io}{https://cua-lite.github.io}}}}
\camerareadytrue
\newcommand{\authornotes}{{\let\thefootnote\relax\footnotetext{$^{\dagger}$Corresponding author: \href{mailto:zhanhui@berkeley.edu}{\texttt{zhanhui@berkeley.edu}}.}}}

\begin{document}
\maketitle
\ifcameraready\authornotes\fi%

\begin{abstract}
Open resources for computer use agents (CUAs) --- environments, datasets, and models --- remain scattered: without a standard, these resources cannot be integrated and scaled; additionally, evaluation, RL, and SFT cannot share one infrastructure. We present \textbf{\Lite{}}, an open platform for developing CUAs, built around three standardized abstractions: \textbf{(1)~Unified environment interface}, under which we integrate $12$ benchmarks across desktop, browser, and mobile, plus lightweight VM-free sandboxes that replicate OSWorld's desktop at much lower cost, providing over $30{,}000$ verifiable tasks. \textbf{(2)~Unified supervised data format}, into which we convert the major public CUA datasets, from GUI grounding and understanding to full rollouts, and record our fresh rollouts from frontier CUAs on these verifiable environments. \textbf{(3)~Model harnesses shared across evaluation, RL, and SFT}, one per model and implemented for $14$ model families. Each harness runs its model in these computer environments, producing the rollouts that evaluation and RL consume, or renders stored supervised data into the model's own training format for SFT. We release reference scores across the integrated benchmarks, and show that RL and SFT through the same harness improve open-weight models. We release our dataset, code, and models, and the platform is open to contributions.
\end{abstract}

\section{Introduction}
\label{sec:intro}

Computer use agents (CUAs) operate desktop, browser, and mobile applications in computer environments. Developing CUAs requires environments with verifiable tasks, supervised data, and models, each with a harness that runs the model in an environment. Open resources exist for all three: environments on desktop~\citep{osworld2024,osworld2_2026,windowsagentarena2024,screenspotpro2025,osworldg2025,cuagym2026,scalecuarl2026,gymanything2026}, browser~\citep{webarena2023,visualwebarena2024,miniwob2018,webvoyager2024,webgym2026}, and mobile~\citep{androidworld2024,androidlab2025,mobileworld2025,mobilegym2026}; supervised datasets, from GUI grounding labels to full rollouts~\citep{mind2web2023,aguvis2024,opencua2025,gui360_2025}; and models, from specialist CUAs~\citep{uitars2025,opencua2025,scalecua2025,evocua2026} and vision--language models (VLMs)~\citep{qwen25vl2025,qwen3vl2025} to API models with their own native computer use tools.

However, these resources remain scattered: each environment has its own interface and runtime (e.g., a virtual machine (VM), a browser, a mobile emulator); each supervised dataset its own format; each model its own action space, prompt format, and rollout code. Without a standard, connecting models to environments means rewriting the same code: an action mapping, context management, and a rollout loop. Therefore, open resources cannot be integrated and scaled; additionally, evaluation, reinforcement learning (RL), and supervised finetuning (SFT) cannot share one infrastructure.

\begin{figure}[t]
\centering
\includegraphics[width=\linewidth]{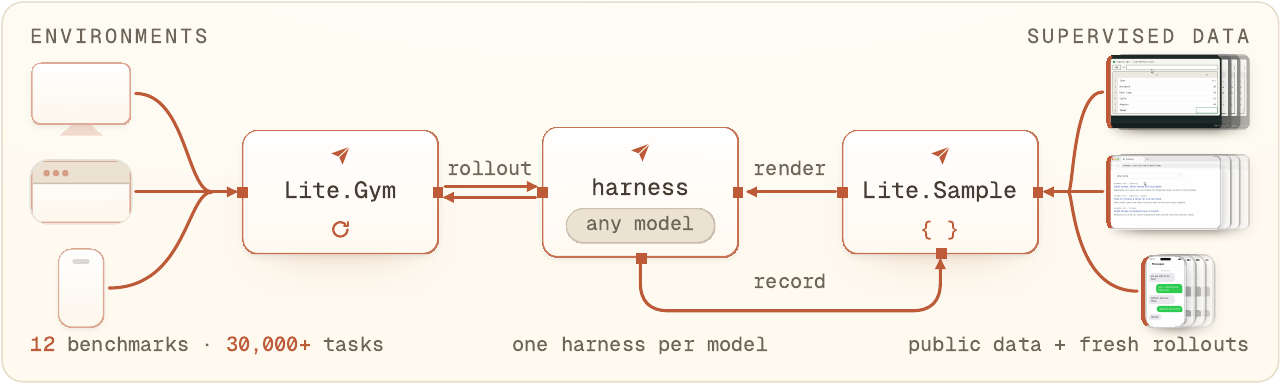}
\caption{\textbf{\Lite{} overview.} One environment interface, \lgym{}, serves every integrated environment, regardless of its runtime. One supervised data format, \lsample{}, stores all supervised data, converted from public datasets or recorded through rollouts. One harness per model connects it to both, so evaluation, RL, and SFT share the same code (\S\ref{sec:harness}).}
\label{fig:overview}
\end{figure}

To address this gap, we present \Lite{}, an open platform for developing CUAs built around three standardized abstractions (Figure~\ref{fig:overview}): \Lite{} unifies environments under one interface, \emph{\lgym{}}; supervised data under one format, \emph{\lsample{}}; and adapts any model to the environment interface and the data format through the model's own harness.

\textbf{(1)~Unified environment interface.} \lgym{} (\S\ref{sec:interface}) exposes an environment as a Gym-style \code{reset}, \code{step}, and \code{close} loop~\citep{gymnasium2025}. The loop standardizes one \emph{observation format} across environments: a screenshot, text, or both. It also standardizes one \emph{GUI action space} per platform, issued as tool calls (e.g., \code{click} and \code{type} on desktop and browser, or \code{tap} and \code{swipe} on mobile). Each environment can add \emph{extra tools} of its own (e.g., \code{bash} in our desktop sandboxes). The loop wraps the environment's own runtime: a VM, a browser, or a mobile emulator. We integrate $12$ benchmarks across desktop, browser, and mobile, and add VM-free desktop sandboxes of our own. These Docker containers replicate OSWorld's desktop at much lower cost and provide over $30{,}000$ verifiable tasks for training (\S\ref{sec:sandboxes}).

\textbf{(2)~Unified supervised data format.} \lsample{} (\S\ref{sec:data}) stores any supervised data, from GUI grounding and understanding to full rollouts, in one standardized format: multi-turn messages that interleave the model's tool calls with the environment's observations. We convert $9$ major public CUA datasets~\citep{mind2web2023,aguvis2024,opencua2025,gui360_2025} and record fresh rollouts from frontier CUAs.

\textbf{(3)~Model harnesses shared across evaluation, RL, and SFT.} Each model has one harness (\S\ref{sec:harness}), the code that adapts the model to the environment interface and the data format. Through its harness, a model runs in every integrated environment, and its rollouts serve both evaluation and RL. The same harness renders any stored \lsample{}, including other models' rollouts, into the model's own training format for SFT. Harnesses exist for $14$ model families, including open-weight models (UI-TARS, Qwen, and others) and API models (GPT, Claude, and Gemini).

We release reference scores for $15$ models on the $12$ integrated benchmarks (\S\ref{sec:evaluation}). We further demonstrate how the rollout data and the verifiable tasks of \Lite{} improve an open-weight computer use agent, by SFTing Qwen3.5-4B on GPT-5.5 rollouts and then training it with GRPO on our verifiable tasks (\S\ref{sec:training}). We release our dataset, code, and models, and welcome contributions.

\section{The \Lite{} Platform}
\label{sec:platform}

In this section we introduce the three standardized abstractions: the environment interface \lgym{} (\S\ref{sec:interface}), under which we integrate existing benchmarks and our VM-free desktop sandboxes with verifiable tasks; the supervised data format \lsample{} (\S\ref{sec:data}), into which we convert public supervised datasets and record fresh rollouts; and the per-model harness (\S\ref{sec:harness}) that runs a model against both. Appendix~\ref{app:related} reviews related work.

\subsection{Environment Interface: \lgym{}}
\label{sec:interface}

\lgym{} exposes every integrated environment through one interface (Figure~\ref{fig:step}), a Gym-style loop~\citep{gymnasium2025} that standardizes one observation format across environments and one GUI action space per platform, plus optional extra tools per environment (Table~\ref{tab:actions}).

\begin{figure}[t]
\centering
\begin{vscodebox}[width=\linewidth,left=5pt,right=5pt]
\begin{pycode}[fontsize=\footnotesize]
env = gym.make("osworld@chrome_030", max_steps=30)   # any integrated environment
obs = await env.reset()                              # obs.image, obs.text

# one step: three GUI actions in one call, one paired result
step = await env.step([make_tool_call("computer", {"actions": [
    {"action": "click", "coordinate": [412, 87]},
    {"action": "type",  "text": "example.com"},
    {"action": "key",   "keys": ["enter"]}]}, call_id="call_0")])
# step.results (one per call id), step.reward, step.terminated, step.truncated

# end to end: the whole rollout, driven by the model's harness
agent  = agents.make("gpt-5.5", env=env)             # resolves the harness
sample = await agent.sample(env)                     # the rollout, as a Lite.Sample
\end{pycode}
\end{vscodebox}
\caption{\textbf{One step and one rollout through the environment interface.} Three GUI actions in one \code{computer} call return one result per \code{call\_id}, so action batching costs one round trip.}
\label{fig:step}
\end{figure}

\textbf{GUI actions as tool calls.} \Lite{} standardizes the GUI actions of each platform and exposes them through one tool per platform: \code{click}, \code{type}, and \code{key} through \code{computer} tool on desktop and browser; \code{tap}, \code{swipe}, and \code{system\_button} through \code{mobile} tool on mobile (Table~\ref{tab:actions}). The model's harness maps its raw actions into that space (\S\ref{sec:harness}), so a model built for one action vocabulary runs on any integrated environment. Each tool call may carry a sequence of actions, so a model can plan several steps in one turn with \emph{action batching} (Figure~\ref{fig:step}). A GUI action's coordinates are normalized to $[0,1000]$ on every platform.

\textbf{Extra tools per environment.} Beyond the GUI actions, an environment optionally exposes tools of its own: \code{bash} on Linux desktops, \code{goto} and \code{back} on browser, and \code{open\_app} on mobile (Table~\ref{tab:actions}). A model issues these tools alongside the platform's GUI actions.

\textbf{Observations as tool results.} A GUI action returns a screenshot, plus any text the environment emits. An extra tool returns its own result, such as a shell command's output.

\begin{table}[t]
\centering
\caption{\textbf{GUI actions and extra tools per platform.} See Appendix~\ref{app:interface} for details.}
\label{tab:actions}
\footnotesize
\setlength{\tabcolsep}{4pt}
\begin{tabular}{@{}l >{\raggedright\arraybackslash}p{0.78\linewidth}@{}}
\toprule
\textbf{Platform} & \textbf{Tool and actions} \\
\midrule
\multicolumn{2}{@{}>{\columncolor{tabgroup}[0pt][0pt]}l@{}}{\textbf{GUI actions}} \\
Desktop, Browser & \code{computer}: \code{click}, \code{mouse\_move}, \code{mouse\_down}, \code{mouse\_up}, \code{drag}, \code{scroll}, \code{type}, \code{key}, \code{key\_down}, \code{key\_up}, \code{hold\_key}, \code{wait}, \code{screenshot}, \code{cursor\_position} \\
Mobile & \code{mobile}: \code{tap}, \code{long\_press}, \code{swipe}, \code{drag}, \code{pinch}, \code{type}, \code{system\_button} \\
Grounding & \code{point}, \code{bbox}: return a coordinate or a box \\
\multicolumn{2}{@{}>{\columncolor{tabgroup}[0pt][0pt]}l@{}}{\textbf{Extra tools}} \\
All platforms & \code{response}, \code{terminate}: finish the task \\
Desktop & \code{bash}: run a shell command \\
Browser & \code{goto}, \code{back}, \code{forward}: navigate \\
 & \code{new\_tab}, \code{switch\_tab}, \code{close\_tab}: manage tabs \\
Mobile & \code{open\_app}: launch an application \\
 & \code{ask\_user}: ask the user a question \\
\bottomrule
\end{tabular}
\end{table}

\subsubsection{Benchmarks and Training Environments}
\label{sec:catalog}

We integrate $12$ benchmarks and $6$ training environments with verifiable tasks across desktop, browser, and mobile (Table~\ref{tab:catalog}). Most environments run unchanged on their own runtimes, with two kinds of exception. Three benchmarks run on third-party implementations: WebVoyager on WebHarbor's offline mirrors of its $15$ websites~\citep{webharbor2026}, and WebArena and MiniWoB++ on BrowserGym~\citep{browsergym2024} (Appendix~\ref{app:benchmarks}). The four \code{Lite.*} environments run on our VM-free desktop sandboxes, described in \S\ref{sec:sandboxes}. 

\begin{table}[t]
\centering
\caption{\textbf{The benchmarks and training environments in \Lite{}.} Each \code{Lite.*} environment runs the tasks of its source (OSWorld~\citep{osworld2024}, SCALECUA~\citep{scalecuarl2026}, CUA-Gym~\citep{cuagym2026}, or Gym-Anything~\citep{gymanything2026}) on our VM-free desktop sandboxes; every other environment runs on its own runtime.}
\label{tab:catalog}
\footnotesize
\setlength{\tabcolsep}{4pt}
\begin{tabular}{@{}l r @{\hskip 18pt} l r@{}}
\toprule
\textbf{Benchmark} & \textbf{\#Eval tasks} & \textbf{Benchmark} & \textbf{\#Eval tasks} \\
\midrule
\multicolumn{2}{@{}l@{}}{\textbf{Grounding}} & \multicolumn{2}{@{}l@{}}{\textbf{Browser}} \\
ScreenSpot-Pro~\citeyearpar{screenspotpro2025} & 1,581 & WebVoyager~\citeyearpar{webvoyager2024} & 643 \\
OSWorld-G~\citeyearpar{osworldg2025} & 564 & WebArena~\citeyearpar{webarena2023} & 812 \\
\multicolumn{2}{@{}l@{}}{\textbf{Desktop}} & MiniWoB++~\citeyearpar{miniwob2018} & 125 \\
OSWorld~\citeyearpar{osworld2024} & 369 & \multicolumn{2}{@{}l@{}}{\textbf{Mobile}} \\
OSWorld 2.1~\citeyearpar{osworld2_2026} & 108 & AndroidWorld~\citeyearpar{androidworld2024} & 116 \\
WindowsAgentArena~\citeyearpar{windowsagentarena2024} & 154 & AndroidLab~\citeyearpar{androidlab2025} & 138 \\
 & & MobileWorld~\citeyearpar{mobileworld2025} & 201 \\
 & & MobileGym~\citeyearpar{mobilegym2026} & 256 \\
\midrule
\textbf{Training environment} & \textbf{\#Train tasks} & \textbf{Training environment} & \textbf{\#Train tasks} \\
\midrule
\multicolumn{2}{@{}l@{}}{\textbf{Desktop}} & \multicolumn{2}{@{}l@{}}{\textbf{Browser}} \\
\losworld{} & 2,429 & WebGym~\citeyearpar{webgym2026} & 292,092 \\
\lscalecua{}~\citeyearpar{scalecuarl2026} & 20,289 & \multicolumn{2}{@{}l@{}}{\textbf{Mobile}} \\
\lcuagym{}~\citeyearpar{cuagym2026} & 10,910 & MobileGym~\citeyearpar{mobilegym2026} & 160 \\
\lcuaworld{}~\citeyearpar{gymanything2026} & 2,479 & & \\
\bottomrule
\end{tabular}
\end{table}

\begin{figure}[t]
\centering
\begin{minipage}[c]{0.62\linewidth}\centering
\footnotesize
\setlength{\tabcolsep}{4pt}
\renewcommand{\arraystretch}{1.25}
\begin{tabular}{@{}l l l@{}}
\toprule
 & \textbf{OSWorld} & \textbf{\losworld{}} \\
\midrule
Runtime & QEMU/KVM VM & Docker container \\
Host & \code{/dev/kvm}, nested virt. & any Docker host \\
Disk & $31$~GB & $6.8$~GB \\
Memory & $3.8$~GB & $0.6$~GB \\
Cold start & $29.3$~s & $21.3$~s \\
CPU when idle & $5.7$~\% & $1.4$~\% \\
\bottomrule
\end{tabular}
\end{minipage}\hfill
\begin{minipage}[c]{0.36\linewidth}\centering
\includegraphics[width=\linewidth]{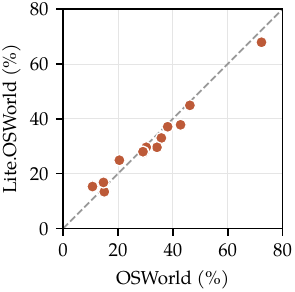}
\end{minipage}
\caption{\textbf{\losworld{} against OSWorld.} Left: cost of one instance, median over $20$ cold starts with both runtimes allocated $4$~GB and $4$ cores; memory (resident usage within that allocation) and CPU are cgroup counters over a $300$~s window with no action sent (Appendix~\ref{app:sandbox}). Right: evaluation scores (\%) of $13$ models. The dashed line marks equal scores. See Table~\ref{tab:parity} for per-model scores.}
\label{fig:sandbox}
\end{figure}

\subsubsection{VM-Free Desktop Sandboxes (\texorpdfstring{\code{Lite.*}}{Lite.*})}
\label{sec:sandboxes}

Most existing desktop environments, such as OSWorld, are distributed as virtual machines. Each instance needs KVM, which shared and cloud hosts often do not expose, and carries a full guest OS, which makes it heavy to run and slow to boot. \textbf{We propose VM-free desktop sandboxes (\code{Lite.*}), containerized Linux desktops that replicate OSWorld's desktop at much lower cost.} The official OSWorld VM ships as a disk image with no build recipe, so we reverse-engineer it. We recover the applications installed in the VM and their configuration, and rebuild the desktop from them as \code{Dockerfile}s. Our rebuilt OSWorld sandbox, \losworld{}, keeps OSWorld's $369$ evaluation tasks, setups, and verifiers unchanged.

\textbf{Cost.} A VM-free sandbox carries no guest OS and shares the host kernel, so it runs on a sixth of OSWorld's memory and boots in three quarters of the time (Figure~\ref{fig:sandbox}, left).

\textbf{Fidelity.} \losworld{} yields evaluation scores close to OSWorld's. We run $13$ models, $12$ open-weight and GPT-5.5, on OSWorld~\citep{osworld2024} and on \losworld{}, over the same $332$ tasks (OSWorld's $369$ minus $29$ infeasible tasks and $8$ that need a Google account) and with identical setup, verifiers, harnesses, and seed (Figure~\ref{fig:sandbox}, right). Scores differ by $2.7$ points on average and by $5.0$ at worst, on EvoCUA-8B, and the two rankings agree with a Spearman correlation of $0.98$ across all $13$ models (Table~\ref{tab:parity}).

\textbf{Training tasks.} Beyond the evaluation tasks they inherit, our sandboxes provide over $30{,}000$ training tasks with verifiers (Table~\ref{tab:catalog}):
\begin{enumerate}[label=\textbf{(\arabic*)}]
\item \losworld{} adds two training splits to OSWorld's $369$ evaluation tasks: $1{,}722$ tasks synthesized from templates and $707$ perturbed from $256$ of them (Appendix~\ref{app:sandbox}).
\item \lscalecua{} runs the $20{,}289$ desktop tasks SCALECUA~\citep{scalecuarl2026} synthesized on that same OSWorld desktop.
\item \lcuagym{} runs the $10{,}910$ desktop and browser tasks of CUA-Gym~\citep{cuagym2026}, across mock sites and real applications.
\item \lcuaworld{} runs $2{,}479$ tasks on the $41$ professional applications of Gym-Anything~\citep{gymanything2026}.
\end{enumerate}

\begin{figure}[t]
\centering
\includegraphics[width=\linewidth]{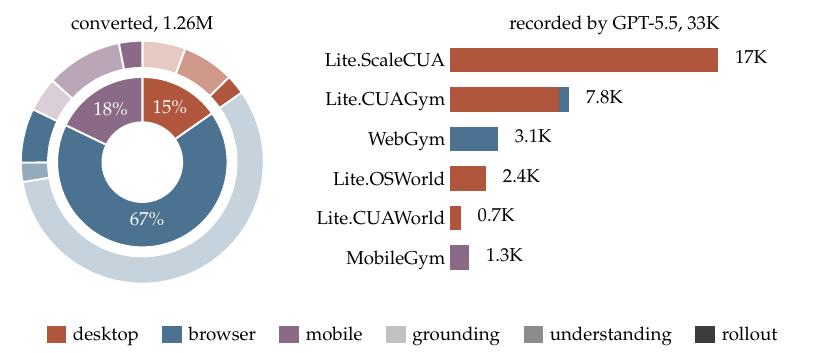}
\caption{\textbf{The stored \lsample{}s.} Left: the $9$ converted datasets, by platform (inner ring) and by task type within each platform (outer ring). Right: GPT-5.5's rollouts with verified reward on our training environments, by platform; Appendix~\ref{app:rollouts} adds those of Qwen3.5-27B and Qwen3.8-27B.}
\label{fig:data}
\end{figure}

\subsection{Supervised Data Format: \lsample{}}
\label{sec:data}

A \lsample{} stores any supervised data (a UI grounding label, UI question understanding, or a full rollout) in one standardized format: multi-turn messages that interleave the model's tool calls with the environment's observations. See Appendix~\ref{app:format} for the schema.

\textbf{Converted datasets.} We convert $9$ major public CUA datasets, covering GUI grounding, understanding, and full rollouts, into $1.26$M \lsample{}s~\citep{mind2web2023,guiodyssey2024,aguvis2024,opencua2025,scalecua2025,gui360_2025} (Figure~\ref{fig:data}, left). See Appendix~\ref{app:converted} for the datasets and the conversion rules.

\textbf{Fresh rollouts.} We record rollouts of GPT-5.5, Qwen3.5-27B, and Qwen3.8-27B on our training environments and keep those with verified reward, $82$K \lsample{}s in all (Figure~\ref{fig:data}, right). Each recorded step stores the model's \code{reasoning\_content}, an \code{action\_description} of the step in one line of natural language, and the \code{tool\_call} it issues, carrying either a batch of GUI actions or one of the environment's extra tools (\S\ref{sec:interface}). Appendix~\ref{app:rollouts} details how we collect them.

\subsection{Model Harness Shared Across Evaluation, RL, and SFT}
\label{sec:harness}

A harness is the scaffolding that makes a model an agent. It turns each observation and the history into the model's input, parses the output back into actions, and drives the loop. \Lite{} has one harness per model, built around an \hpart{adapter} that converts between a \lsample{} and the model's own input and output. For SFT, the adapter renders a stored \lsample{} into the model's own format for training (\S\ref{sec:data}). During a rollout for evaluation or RL, it renders the \lsample{} as it grows and converts the model's reply back into \lgym{} actions. Inside the adapter, \hpart{action\_space} maps the model's raw actions to and from the standardized GUI actions (\S\ref{sec:interface}), and \hpart{protocol} manages the model's context across turns. Modules are shared and inherited across models (Qwen3.5, Qwen3.8, and EvoCUA all build on Qwen3-VL's), so a new harness implements only what differs. Harnesses currently exist for $14$ model families.

\begin{table}[t]\centering
\caption{\textbf{Reference evaluation scores (\%) on all $12$ benchmarks.} Bold: best per benchmark; underline: best open-weight model. WAA: WindowsAgentArena. --: not applicable. Qwen models run with reasoning off and API models at medium effort. See Appendix~\ref{app:eval} for details.}
\label{tab:main}
\footnotesize
\setlength{\tabcolsep}{4.3pt}
\newcommand{\rot}[1]{\adjustbox{angle=45,lap=\width-1.1em}{#1}}
\begin{tabular}{@{}l cc @{\hskip 7pt} ccc @{\hskip 7pt} ccc @{\hskip 7pt} cccc@{}}
\toprule
\textbf{Model} & \rot{ScreenSpot-Pro} & \rot{OSWorld-G} & \rot{OSWorld} & \rot{OSWorld 2.1} & \rot{WAA} & \rot{WebVoyager} & \rot{WebArena} & \rot{MiniWoB++} & \rot{AndroidWorld} & \rot{AndroidLab} & \rot{MobileWorld} & \rot{MobileGym} \\
& \multicolumn{2}{c}{\scriptsize Grounding} & \multicolumn{3}{c}{\scriptsize Desktop} & \multicolumn{3}{c}{\scriptsize Browser} & \multicolumn{4}{c}{\scriptsize Mobile} \\
\cmidrule(lr){2-3}\cmidrule(lr){4-6}\cmidrule(lr){7-9}\cmidrule(lr){10-13}
\multicolumn{13}{@{}>{\columncolor{tabgroup}[0pt][0pt]}l@{}}{\textbf{Specialist CUAs}} \\
UI-TARS-1.5-7B    & 49.8 & 60.6 & 27.5 & 0.5 & 16.7 & 46.2 & 9.1 & 18.4 & 19.8 & 36.2 & 11.1 & 13.3   \\
EvoCUA-8B         & 43.5 & 57.3 & 35.6 & 0.7 & 16.6 & 5.8 & 10.8 & 56.0 & -- & -- & -- & --   \\
MAI-UI-8B         & 62.2 & 70.0 & -- & -- & -- & -- & -- & -- & 61.2 & 49.3 & 29.9 & 25.8   \\
UI-Venus-2-9B     & 71.9 & 78.2 & 50.5 & 1.4 & \underline{41.3} & \underline{72.9} & \underline{30.7} & 59.2 & 62.1 & 47.8 & \underline{59.0} & 34.4   \\
\multicolumn{13}{@{}>{\columncolor{tabgroup}[0pt][0pt]}l@{}}{\textbf{General VLMs}} \\
Qwen3-VL-4B       & 58.3 & 65.5 & 23.6 & 0.9 & 9.5 & 20.6 & 8.7 & 47.2 & 43.1 & 36.2 & 6.8 & 13.7   \\
Qwen3-VL-8B       & 55.2 & 64.3 & 28.1 & 0.3 & 14.5 & 29.3 & 7.1 & 53.6 & 51.7 & 39.9 & 8.5 & 16.0   \\
Qwen3-VL-32B      & 57.7 & 72.0 & 34.3 & 1.2 & 24.0 & 42.8 & 14.6 & 60.8 & 62.1 & 36.2 & 17.9 & 23.4   \\
Qwen3.5-4B        & 57.7 & 62.9 & 27.0 & 0.7 & 11.6 & 12.2 & 11.3 & 28.0 & 60.3 & 34.8 & 11.1 & 17.6   \\
Qwen3.5-9B        & 58.1 & 68.2 & 38.6 & 0.4 & 13.1 & 35.0 & 15.4 & 50.4 & 62.1 & 36.2 & 10.3 & 18.0   \\
Qwen3.5-27B       & 69.5 & 74.3 & 48.9 & \underline{2.8} & 25.4 & 51.7 & 21.9 & \underline{68.8} & \underline{65.1} & 40.6 & 15.4 & 26.6   \\
Qwen3.8-27B       & \underline{75.4} & \underline{79.6} & \underline{59.2} & 2.4 & 37.0 & 55.4 & 24.5 & 64.8 & 64.7 & \underline{51.4} & 32.5 & \underline{40.2}   \\
\multicolumn{13}{@{}>{\columncolor{tabgroup}[0pt][0pt]}l@{}}{\textbf{API models}} \\
Claude Opus 5     & 85.3 & 89.2 & \textbf{78.7} & \textbf{63.1} & \textbf{72.4} & \textbf{90.4} & 51.7 & \textbf{91.2} & 81.0 & \textbf{62.3} & 78.6 & 70.2  \\
GPT-5.5           & 74.5 & 90.4 & 71.1 & 38.6 & 66.6 & 78.4 & 47.1 & 84.8 & 78.4 & 60.9 & 80.3 & 53.9  \\
GPT-5.6 Sol       & 82.5 & 91.2 & 73.3 & 43.9 & 70.0 & 79.8 & \textbf{53.3} & 82.4 & 75.9 & 60.9 & \textbf{86.3} & 64.1  \\
GPT-6 Astra       & \textbf{93.9} & \textbf{92.5} & 75.1 & 63.0 & 63.4 & 84.3 & \textbf{53.3} & 87.2 & \textbf{84.5} & 60.9 & 85.5 & \textbf{80.9}  \\
\bottomrule
\end{tabular}
\end{table}

\section{Evaluation}
\label{sec:evaluation}

In this section we run each model on the benchmarks it supports and release the resulting reference scores (Table~\ref{tab:main}). \textit{Most previously reported scores come with no runnable evaluation script, so we do not claim to reproduce them; instead, we offer these scores as a reference that can be rerun from the \Lite{} configuration we release}.

\paragraph{Models.} We run $15$ models: specialist CUAs (UI-TARS-1.5, EvoCUA, MAI-UI, UI-Venus-2), general VLMs (Qwen3-VL at $4$B--$32$B, Qwen3.5 at $4$B--$27$B, Qwen3.8-27B), and API models (Claude Opus 5, GPT-5.5, GPT-5.6 Sol, GPT-6 Astra). Each runs through its own harness (\S\ref{sec:harness}), with the settings of Table~\ref{tab:harness}. By default, open-weight models run with reasoning off and API models at medium reasoning effort. The one exception is UI-Venus-2, which by its official default runs with reasoning on for every task other than grounding.

\paragraph{Benchmarks.} We run the $12$ benchmarks of Table~\ref{tab:catalog} across grounding, desktop, browser, and mobile. Most scores come from one rollout per task under one fixed seed, with greedy decoding and each benchmark's own step budget. See Appendix~\ref{app:eval} for the tasks we evaluate.

\paragraph{Results.} At the same default reasoning effort (medium), GPT-6 Astra and Claude Opus 5 score highest, Astra on grounding and Opus 5 on most desktop and browser benchmarks (Table~\ref{tab:main}). Among open-weight models, Qwen3.8-27B and UI-Venus-2-9B score highest, and each Qwen family improves with model size on average.

\begin{figure}[t]
\centering
\includegraphics[width=\linewidth]{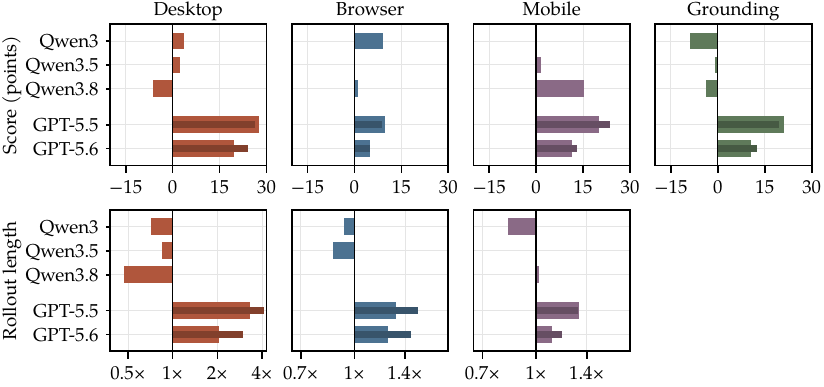}
\caption{\textbf{Effect of reasoning on score and rollout length.} Top: the change in score. Bottom: the ratio of rollout lengths. For Qwen models, each bar compares reasoning on with reasoning off; for GPT models, \code{medium} (wide bar) or \code{xhigh} (narrow, darker bar) reasoning effort with \code{none}. Each platform averages its benchmarks. The Qwen3 and Qwen3.5 rows average Qwen3-VL (4B--32B) and Qwen3.5 (4B--27B), shown per model in Figure~\ref{fig:reasoning-models}.}
\label{fig:reasoning}
\end{figure}

\paragraph{Reasoning ablation.} We rerun Qwen3-VL, Qwen3.5, and Qwen3.8 with reasoning on and off (default), and GPT-5.5 and GPT-5.6 Sol at \code{none}, \code{medium} (default), and \code{xhigh} reasoning effort (Figure~\ref{fig:reasoning}, Table~\ref{tab:eval-full}). \textbf{Reasoning helps the GPT models far more than the Qwen models.} The GPT models gain on every platform from \code{none} to \code{medium}, and little more at \code{xhigh}, and their rollouts grow longer as they verify their work before terminating. For the Qwen models, reasoning lowers every model's grounding average and mostly shortens their rollouts, but its effect on their scores on the other platforms varies. For example, Qwen3.8-27B gains $15$ points on mobile but loses $6.3$ on desktop, where its rollouts are $53\%$ shorter than without reasoning.

\section{Training}
\label{sec:training}

In this section, we demonstrate how the rollout data and the verifiable tasks of \Lite{} improve an open-weight computer use agent through SFT and RL. Specifically, we first SFT Qwen3.5-4B on the GPT-5.5 rollouts of \S\ref{sec:data} (\S\ref{sec:sft}), and then further train the SFTed model with GRPO~\citep{deepseekmath2024} on the verifiable tasks of \S\ref{sec:catalog} (\S\ref{sec:rl}). Every SFT and RL run uses a single node of 8 A100 GPUs, 96 vCPU, 1.8 TB RAM.

\subsection{SFT}
\label{sec:sft}

\paragraph{Setup.} We SFT Qwen3.5-4B on GPT-5.5 rollouts collected from the verifiable training tasks (\S\ref{sec:data}) that pass the verifier, and evaluate on held-out tasks. For desktop, we train on $5{,}000$ \lsample{}s from \lscalecua{} and evaluate on \losworld{}'s $332$ evaluation tasks (OSWorld's $369$, excluding the $29$ infeasible tasks and the $8$ that need a Google account). For browser, we train on about $2{,}000$ \lsample{}s from WebGym and evaluate on $128$ held-out WebVoyager tasks. For mobile, we train on about $1{,}000$ \lsample{}s from MobileGym, whose $160$ training tasks are templates that each generate many instances, and evaluate on its own $256$ held-out tasks. We train Qwen3.5-4B with its official computer use harness (\S\ref{sec:harness}), with an image history of one screenshot ($1920\times1080$ on desktop, $1280\times720$ on browser, $720\times1600$ on mobile) and the full text history. We evaluate every model three times at temperature $0$ and report the mean success rate (Appendix~\ref{app:training}).

\paragraph{Results.} SFT on the GPT-5.5 rollouts raises Qwen3.5-4B from $22.6$ to $44.8$ on desktop, from $4.4$ to $60.0$ on browser, and from $14.6$ to $26.8$ on mobile (Table~\ref{tab:sft}). As an ablation, we run SFT on these rollouts without their reasoning (\S\ref{sec:data}), which costs $7.6$, $11.3$, and $5.4$ points respectively, while Qwen3.5-4B's own reasoning does not always help before SFT.

\begin{figure}[t]
\centering
\includegraphics[width=\linewidth]{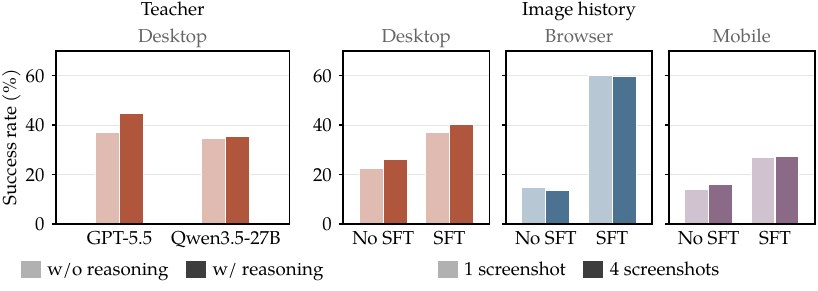}
\caption{\textbf{SFT ablations of Qwen3.5-4B.} Left: desktop evaluation scores (\%) after SFT on each teacher's rollouts, with and without reasoning. Right: evaluation scores (\%) with the one or four most recent screenshots in context, without SFT and after SFT on GPT-5.5 rollouts, both with reasoning. Desktop screenshots are $1920\times1080$ on the left and $1280\times720$ on the right, as in Table~\ref{tab:sft-full}.}
\label{fig:sft-ablation}
\end{figure}

\paragraph{Ablations.} We also ablate two choices of the SFT setup: the teacher that provides the rollouts and the image history (Table~\ref{tab:sft-full} gives the exact scores).
\begin{enumerate}[label=\textbf{(\arabic*)}]
\item \textbf{Teacher} (Figure~\ref{fig:sft-ablation}, left). We ablate the model whose rollouts Qwen3.5-4B is SFTed on, replacing GPT-5.5 with Qwen3.5-27B on desktop. \textbf{GPT-5.5 is the stronger teacher, and most of its lead comes from reasoning}. Without reasoning, Qwen3.5-27B's rollouts train a model close to GPT-5.5's, and adding its reasoning helps little.
\item \textbf{Image history} (Figure~\ref{fig:sft-ablation}, right). We ablate how many of the most recent screenshots Qwen3.5-4B keeps in its context, one or four. \textbf{The latest screenshot alone does about as well as four}, both before and after SFT.
\end{enumerate}

\begin{table}[t]\centering
\caption{\textbf{Evaluation scores (\%) of Qwen3.5-4B before and after SFT on GPT-5.5 rollouts.} The rows without SFT differ by whether Qwen3.5-4B's own reasoning is on, the SFT rows by whether the GPT-5.5 rollouts used for SFT contain reasoning. The rows without SFT use the configuration of SFT (one screenshot) for fair comparison rather than the official configuration of Table~\ref{tab:main} (up to four screenshots), so their scores differ from Table~\ref{tab:main}.}
\label{tab:sft}
\footnotesize
\setlength{\tabcolsep}{8pt}
\begin{tabular}{@{}l ccc@{}}
\toprule
 & \textbf{Desktop} & \textbf{Browser} & \textbf{Mobile} \\
 & \losworld{} & WebVoyager & MobileGym \\
\midrule
\multicolumn{4}{@{}>{\columncolor{tabgroup}[0pt][0pt]}l@{}}{\textbf{Without SFT}} \\
w/o reasoning & 22.6 & 4.4 & 14.6 \\
w/\phantom{o} reasoning & 27.1 & 14.8 & 13.9 \\
\multicolumn{4}{@{}>{\columncolor{tabgroup}[0pt][0pt]}l@{}}{\textbf{SFT on GPT-5.5 rollouts}} \\
w/o reasoning          & 37.2 & 48.7 & 21.4 \\
w/\phantom{o} reasoning & \textbf{44.8} & \textbf{60.0} & \textbf{26.8} \\
\bottomrule
\end{tabular}
\end{table}

\subsection{RL}
\label{sec:rl}

\paragraph{Setup.} We train the SFTed Qwen3.5-4B with GRPO on verifiable tasks (\S\ref{sec:catalog}) and evaluate on held-out tasks. For desktop, we train on the $814$ LibreOffice tasks (Impress, Calc, Writer) that \losworld{} synthesizes, and evaluate on the $117$ LibreOffice tasks of the official OSWorld evaluation set. For browser, we use the $622$ WebVoyager tasks that do not modify their site, on WebHarbor's offline mirrors, and train on $494$ of them and evaluate on the other $128$, the held-out tasks of \S\ref{sec:sft}. For mobile, we drop the hardest level of MobileGym's evaluation split and group the rest into $37$ task families. We train on one task from each family and evaluate on the $52$ remaining tasks of those families, with a shaped reward that gives partial credit for passed goal checks. We run five seeds per platform (Appendix~\ref{app:training}).

\begin{figure}[t]
\centering
\includegraphics[width=\linewidth]{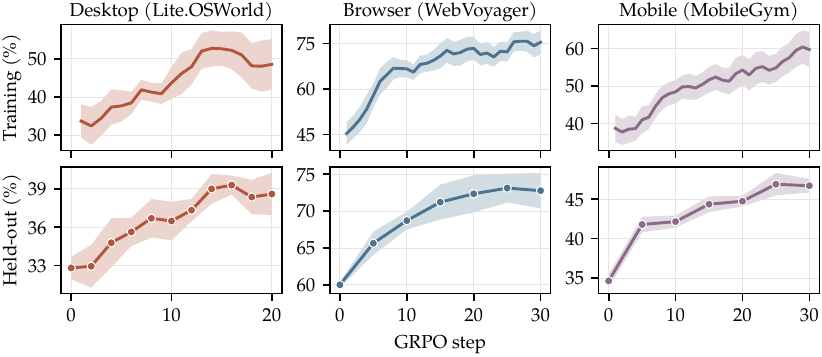}
\caption{\textbf{GRPO from the SFTed Qwen3.5-4B.} Top: training reward per GRPO step, smoothed over five GRPO steps. Bottom: held-out evaluation score. Both rows show the mean and standard error over five seeds. Mobile evaluates on a subset of Table~\ref{tab:sft}'s held-out tasks, and desktop on OSWorld's LibreOffice tasks.}
\label{fig:rl}
\end{figure}

\paragraph{Results.} GRPO further improves the SFTed Qwen3.5-4B on every platform (Figure~\ref{fig:rl}). On held-out tasks, the evaluation score rises from $32.8$ to $38.6$ on desktop, the success rate from $60.0$ to $72.8$ on browser, and the shaped reward from $34.6$ to $46.7$ on mobile.

\paragraph{Choosing training tasks.} \textbf{The obvious training tasks do not always suit RL.} On desktop, before the runs in Figure~\ref{fig:rl}, we tried GRPO on the official \lscalecua{} RL split but found it did not raise the held-out score. We hypothesize that SFT on GPT-5.5's \lscalecua{} rollouts had already saturated the model on those tasks. The desktop runs in Figure~\ref{fig:rl} therefore use \losworld{}'s synthesized LibreOffice tasks, one application suite so that a run fits on one node. On browser, training on live websites is prone to IP blocking, so we instead train on WebHarbor's offline mirrors, using the WebVoyager tasks outside our held-out set. On mobile, MobileGym's official training split is much easier than its evaluation split, so we draw our training and held-out tasks from it.

\section{Limitations and Conclusion}
\label{sec:conclusion}

\textbf{Limitations.} \Lite{} standardizes how environments, data, and models connect, but it also inherits their limitations. Open mobile environments with verifiable training tasks are scarce, so MobileGym is our only mobile training environment. WebVoyager runs on offline mirrors of its websites, so its scores do not measure the live sites. We do not fully audit the training verifiers, and a model judge scores some tasks in \lcuagym{} and \lcuaworld{}, so SFT data and RL rewards can include rollouts that did not solve their task. Our VM-free desktop sandboxes cover only Linux desktops, so Windows environments such as WindowsAgentArena still need a virtual machine, and mobile environments such as AndroidWorld still need an emulator. In \S\ref{sec:training}, the desktop and mobile models are evaluated on the applications they train on, so their gains do not show transfer to unseen applications. Most previously reported scores come with no runnable evaluation script, so we do not claim to reproduce them and offer ours only as reference scores. Our harnesses implement only each model's own GUI loop, not richer harnesses such as Codex and Claude Code.

\textbf{Conclusion.} \Lite{} gives computer use agent development one standard: an environment interface, a supervised data format, and per-model harnesses shared across evaluation, RL, and SFT. On that standard we integrate $12$ benchmarks and VM-free desktop sandboxes with over $30{,}000$ verifiable tasks, convert the major public CUA datasets, and record fresh rollouts. \Lite{} stays open to contributions, and every number in this paper ships with the command that produces it.

\IfFileExists{\jobname.bbl}{\bibliography{main}}{
}

\appendix
\clearpage
\counterwithin{figure}{section}
\counterwithin{table}{section}
\section{Related Work}
\label{app:related}

\textbf{Environments} for CUAs span desktop~\citep{osworld2024,osworld2_2026,windowsagentarena2024,screenspotpro2025,osworldg2025,cuagym2026,scalecuarl2026,gymanything2026}, browser~\citep{miniwob2018,webarena2023,visualwebarena2024,webvoyager2024,webgym2026}, and mobile~\citep{androidenv2021,androidworld2024,androidlab2025,mobileworld2025,mobilegym2026}; each defines its own action space and runtime, and most desktop benchmarks are distributed as virtual machines. \textbf{Datasets} range from GUI grounding and understanding labels to full rollouts, each in its own format~\citep{mind2web2023,guiodyssey2024,aguvis2024,opencua2025,scalecua2025,gui360_2025}. \textbf{Models} range from specialist CUAs~\citep{uitars2025,opencua2025,scalecua2025,maiui2025,evocua2026,fara2025,uivenus2026} to VLMs~\citep{qwen25vl2025,qwen3vl2025} and API models with native computer use tools, each with its own action space and prompt format~\citep{guiagentsurvey2024}. \textbf{Systems and infrastructure.} OpenHands targets software engineering~\citep{openhands2025}, BrowserGym and AgentLab the evaluation of web agents~\citep{browsergym2024}, and ClawGUI the training and deployment of mobile agents~\citep{clawgui2026}; AgentStudio is a toolkit of environments, annotation tools, and benchmarks for virtual agents~\citep{agentstudio2025}, and AgentBoard focuses on analytical evaluation~\citep{agentboard2024}. GEM and AgentGym-RL provide gym-style interfaces for text agents~\citep{gem2026,agentgymrl2026}; ComputerRL, SCALECUA, CUA-Gym, MobileGym, and WebGym scale environments or verified tasks for one platform~\citep{computerrl2026,scalecuarl2026,cuagym2026,mobilegym2026,webgym2026}; OpenEnv standardizes environment packaging~\citep{openenv2025}. Each of these systems covers one platform, or one of the three abstractions. \Lite{} provides all three across desktop, browser, and mobile, and one harness per model serves evaluation, RL, and SFT alike.

\section{Environment Interface (\lgym{})}
\label{app:interface}
Table~\ref{tab:gui-actions} lists every GUI action with its arguments, and Table~\ref{tab:extra-tools} the extra tools, their arguments, and the environments that expose them.

\begingroup\renewcommand{\code}[1]{\texttt{#1}}
\begin{table}[h]\centering
\caption{\textbf{GUI actions and their arguments.} Coordinates are normalized to $[0,1000]$ (\S\ref{sec:interface}).}
\label{tab:gui-actions}
\footnotesize
\setlength{\tabcolsep}{4pt}
\begin{tabular}{@{}l l@{}}
\toprule
\textbf{Action} & \textbf{Arguments} \\
\midrule
\multicolumn{2}{@{}>{\columncolor{tabgroup}[0pt][0pt]}l@{}}{\textbf{computer} (desktop, browser)} \\
\code{click}, \code{mouse\_down}, \code{mouse\_up} & \code{coordinate}, \code{button}, \code{clicks}$\in\{1,2,3\}$ \\
\code{mouse\_move} & \code{coordinate} \\
\code{drag} & \code{start\_coordinate}, \code{coordinate}, \code{button} \\
\code{scroll} & \code{direction}, \code{amount}, \code{coordinate} \\
\code{type} & \code{text}, \code{press\_enter} \\
\code{key}, \code{key\_down}, \code{key\_up}, \code{hold\_key} & \code{keys}, \code{duration} \\
\code{wait} & \code{duration} \\
\code{screenshot}, \code{cursor\_position} & none \\
\multicolumn{2}{@{}>{\columncolor{tabgroup}[0pt][0pt]}l@{}}{\textbf{mobile} (mobile)} \\
\code{tap}, \code{long\_press} & \code{coordinate}, \code{clicks}, \code{duration} \\
\code{swipe}, \code{drag}, \code{pinch} & \code{start\_coordinate}, \code{coordinate}, \code{direction} \\
\code{type} & \code{text} \\
\code{system\_button} & \code{button} \\
\code{wait} & \code{duration} \\
\code{screenshot} & none \\
\multicolumn{2}{@{}>{\columncolor{tabgroup}[0pt][0pt]}l@{}}{\textbf{Grounding}} \\
\code{point}, \code{bbox} & \code{coordinate}, \code{box} \\
\bottomrule
\end{tabular}
\end{table}\endgroup

\begingroup\renewcommand{\code}[1]{\texttt{#1}}
\begin{table}[h]\centering
\caption{\textbf{Extra tools per platform.}}
\label{tab:extra-tools}
\footnotesize
\setlength{\tabcolsep}{4pt}
\begin{tabular}{@{}l l >{\raggedright\arraybackslash}p{0.46\linewidth}@{}}
\toprule
\textbf{Tool} & \textbf{Arguments} & \textbf{Environments} \\
\midrule
\multicolumn{3}{@{}>{\columncolor{tabgroup}[0pt][0pt]}l@{}}{\textbf{All platforms}} \\
\code{response} & \code{text} & every non-grounding environment \\
\code{terminate} & \code{status}, \code{reason} & every non-grounding environment \\
\multicolumn{3}{@{}>{\columncolor{tabgroup}[0pt][0pt]}l@{}}{\textbf{Desktop}} \\
\code{bash} & \code{command} & \losworld{}, \lscalecua{}, \lcuagym{}, \lcuaworld{} \\
\multicolumn{3}{@{}>{\columncolor{tabgroup}[0pt][0pt]}l@{}}{\textbf{Browser}} \\
\code{goto} & \code{url} & WebGym, WebArena, MiniWoB++ \\
\code{back} & none & WebGym, WebArena, MiniWoB++, WebVoyager \\
\code{forward}, \code{new\_tab}, \code{close\_tab} & none & WebArena, MiniWoB++ \\
\code{switch\_tab} & \code{index} & WebArena, MiniWoB++ \\
\multicolumn{3}{@{}>{\columncolor{tabgroup}[0pt][0pt]}l@{}}{\textbf{Mobile}} \\
\code{open\_app} & \code{app\_name} & AndroidWorld, AndroidLab, MobileWorld, MobileGym \\
\code{ask\_user} & \code{question} & MobileWorld \\
\bottomrule
\end{tabular}
\end{table}\endgroup

\section{Benchmark Implementations}
\label{app:benchmarks}
We wrap each benchmark's official implementation, with its tasks, runtime, and evaluator, except for three browser benchmarks that run on a third-party implementation.

\paragraph{WebVoyager.} WebVoyager's tasks run on WebHarbor's offline mirrors of its $15$ websites~\citep{webharbor2026}, served from one container whose databases reset when it starts, so its scores measure the mirrors rather than the live sites. A GPT-4.1 judge scores the last three screenshots and the final answer of each rollout.

\paragraph{WebArena and MiniWoB++.} Both run on their BrowserGym implementations~\citep{browsergym2024}, WebArena against its self-hosted website containers and MiniWoB++ against locally served pages. BrowserGym's fuzzy-match judge for WebArena runs on GPT-4.1 instead of the retired GPT-4 Turbo model it hard-codes.

\section{VM-Free Desktop Sandboxes}
\label{app:sandbox}
\begin{figure}[p]\centering
\includegraphics[width=\linewidth]{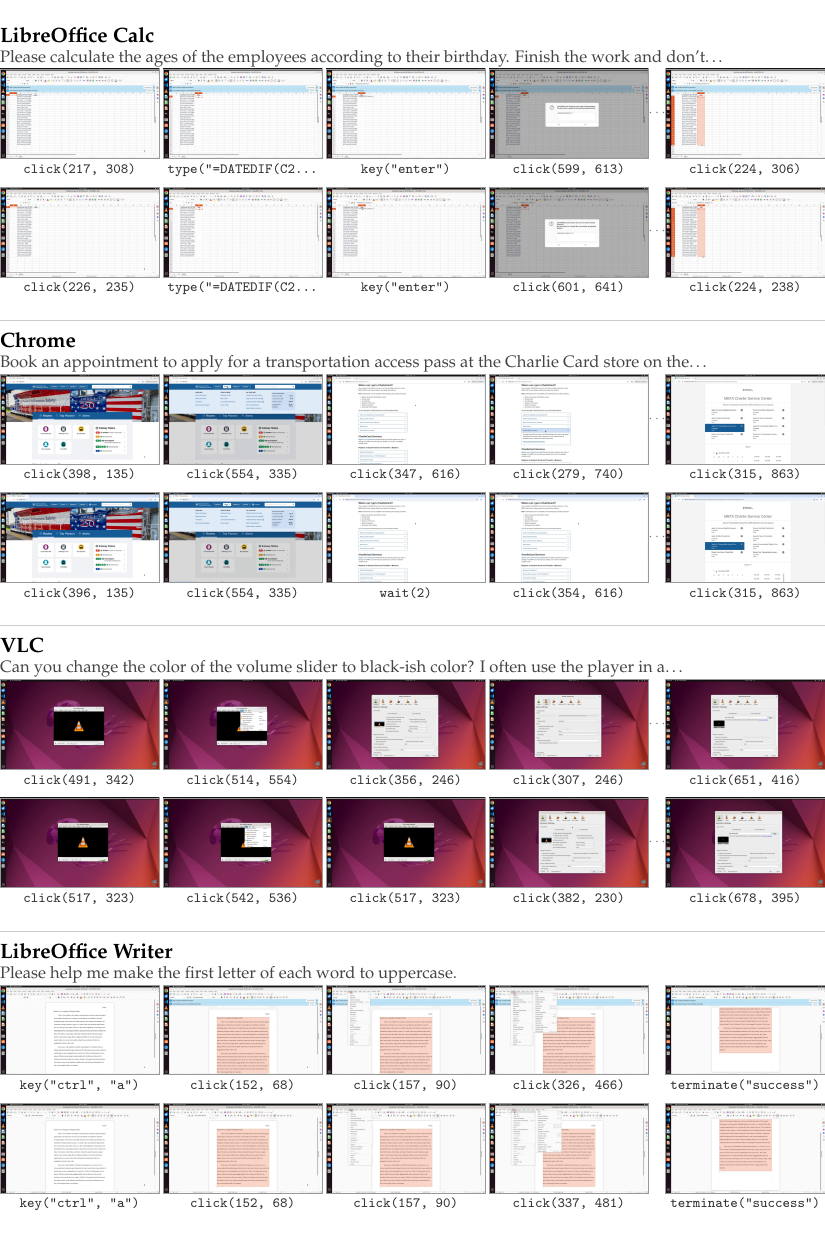}
\caption{\textbf{Four tasks, step by step, on both runtimes.} Each block is one task, the OSWorld VM on the upper row and the \losworld{} container on the lower. A row gives a rollout's first four steps and its last, with $\cdots$ for the steps between and, under each panel, the action GPT-5.5 took from that screen.}
\label{fig:traj}
\end{figure}

\paragraph{Fidelity.} Table~\ref{tab:parity} gives the per-model scores behind Figure~\ref{fig:sandbox} (right), comparing OSWorld and \losworld{} on the same $332$ tasks (OSWorld's $369$, excluding the $29$ infeasible tasks and the $8$ that need a Google account; Appendix~\ref{app:eval}) under the same setup, verifiers, and harness. Figure~\ref{fig:traj} shows four of those tasks (in LibreOffice Calc, Chrome, VLC, and LibreOffice Writer) step by step on both runtimes.

\begin{table}[t]\centering
\caption{\textbf{Evaluation scores (\%) on OSWorld and \losworld{}.} Both runtimes run in one parity campaign with the same settings, separate from the reference run of Table~\ref{tab:main}.}
\label{tab:parity}
\footnotesize
\setlength{\tabcolsep}{6pt}
\begin{tabular}{@{}l ccc@{}}
\toprule
\textbf{Model} & \textbf{OSWorld} & \textbf{\losworld{}} & $\bm{\Delta}$ \\
\midrule
GPT-5.5        & 72.3 & 67.9 & $-4.4$ \\
Qwen3.5-27B    & 46.2 & 44.9 & $-1.3$ \\
EvoCUA-8B      & 42.8 & 37.8 & $-5.0$ \\
Qwen3.5-9B     & 38.1 & 37.1 & $-1.0$ \\
Qwen3-VL-32B   & 35.8 & 33.0 & $-2.8$ \\
Qwen3-VL-8B    & 34.2 & 29.6 & $-4.6$ \\
Qwen3.5-4B     & 30.3 & 29.6 & $-0.7$ \\
UI-TARS-1.5-7B & 29.4 & 27.7 & $-1.7$ \\
Qwen3-VL-4B    & 29.1 & 28.0 & $-1.1$ \\
Qwen3-VL-2B    & 20.5 & 24.9 & $+4.4$ \\
ScaleCUA-7B    & 15.0 & 13.4 & $-1.6$ \\
UI-TARS-7B-DPO & 14.7 & 16.8 & $+2.1$ \\
Qwen3.5-2B     & 10.7 & 15.3 & $+4.6$ \\
\bottomrule
\end{tabular}
\end{table}

\paragraph{Cost.} Figure~\ref{fig:sandbox} (left) measures one instance of each runtime under a matched allocation. OSWorld distributes its VM with $4$~GB of guest memory and $4$ cores, so we cap the \losworld{} container at the same $4$~GB and $4$ cores for measurement.

\begin{itemize}[topsep=2pt,itemsep=1pt,parsep=0pt,leftmargin=1.2em]
\item \textbf{Cold start}: the wall time from \code{gym.make} to the first observation \code{reset} returns, which covers the boot, the task's setup, and the first screenshot. Setup differs from task to task, so we alternate two \code{os}-domain tasks rather than repeat one, and both runtimes run that same pair, which leaves setup equal on the two sides. We report the median of $20$ cold starts per runtime, interleaved so that host drift reaches both.
\item \textbf{Memory}: resident memory, read from the instance's cgroup as \code{memory.current} minus \code{inactive\_file}, at the end of a $300$~s idle window that opens $60$~s after \code{reset} returns, with no action sent. Both numbers are usage rather than the cap. A container takes what it touches, and a VM holds the guest RAM its host must provide.
\item \textbf{Idle CPU}: the cgroup's cumulative CPU time (\code{cpu.stat}'s \code{usage\_usec}) differenced across that same window and expressed as a share of one core, rather than a sampled rate, because one desktop repaint moves a sampled percentage by two orders of magnitude.
\item \textbf{Disk}: the image size, plus OSWorld's \code{qcow2}.
\end{itemize}

\paragraph{Training tasks.} \lscalecua{}, \lcuagym{}, and \lcuaworld{} inherit the task sets of their upstream releases, with the bugs we found fixed. \losworld{} adds two training splits to OSWorld's $369$ evaluation tasks (Table~\ref{tab:losworld-train}):
\begin{itemize}[topsep=2pt,itemsep=1pt,parsep=0pt,leftmargin=1.2em]
\item \textbf{Synthesized} ($1{,}722$ tasks): we generate each task from a template, and the templates of each application cover the operations of its OSWorld evaluation tasks. The tasks do not use OSWorld's files. Most generate their own in the sandbox or copy in public ones (e.g., Wikimedia Commons photos and Project Gutenberg books), and the rest start from the application alone. Only the Thunderbird tasks reuse OSWorld's mail profile.
\item \textbf{Perturbed} ($707$ tasks): we perturb $256$ of OSWorld's evaluation tasks, keeping each task's files and setup and changing the request.
\end{itemize}
Both splits mostly reuse OSWorld's evaluator functions as verifiers, and we check that each task's oracle solution scores $1$. We train on the synthesized tasks (\S\ref{sec:rl}). The perturbed tasks start from the same state as evaluation tasks, so they are useful for debugging and analysis; we do not train on them and do not recommend it, as it would overfit to OSWorld.

\begin{table}[h]\centering
\caption{\textbf{\losworld{} tasks per application.}}
\label{tab:losworld-train}
\footnotesize
\setlength{\tabcolsep}{8pt}
\begin{tabular}{@{}l rr r@{}}
\toprule
 & \multicolumn{2}{c}{\textbf{Training}} & \\
\cmidrule(lr){2-3}
\textbf{Application} & Synthesized & Perturbed & \textbf{Evaluation} \\
\midrule
Chrome              & 132 & 91 & 46 \\
GIMP                & 73 & 42 & 26 \\
LibreOffice Calc    & 278 & 106 & 47 \\
LibreOffice Impress & 287 & 108 & 47 \\
LibreOffice Writer  & 249 & 43 & 23 \\
Multiple apps       & 379 & 138 & 101 \\
OS                  & 147 & 34 & 24 \\
Thunderbird         & 48 & 32 & 15 \\
VLC                 & 37 & 44 & 17 \\
VS Code             & 92 & 69 & 23 \\
\midrule
Total               & 1,722 & 707 & 369 \\
\bottomrule
\end{tabular}
\end{table}

\clearpage%
\section{Datasets (\lsample{})}
\label{app:data}

\subsection{Format}
\label{app:format}

A \lsample{} stores supervised data (a UI grounding label, UI question understanding, or a full rollout) in three fields: the images it refers to, the conversation, and the metadata that labels it (Figure~\ref{fig:schema}). Each message is tagged with its role (user, assistant, or tool), each content part with its kind (text, image, or action description), and the reasoning has its own \code{reasoning\_content} field. Because every part is tagged, a harness can find a part by its type instead of parsing text. For example, it can drop the reasoning, keep only the most recent screenshots, or replace past steps with their action descriptions. Figure~\ref{fig:sample} shows one \lsample{} of each kind.

\begin{figure}[h]
\begin{vscodebox}[width=\linewidth,left=5pt,right=5pt,before skip=1pt,after skip=1pt]
\begin{pycode}[escapeinside=??,texcomments=false]
@dataclass
class LiteSample:                    # one label, answer, or rollout
    images: list[Image]              # what the messages refer to, deduplicated
    messages: list[LiteMessage]      # user, assistant and tool turns
    metadata: LiteCUAMetadata

class LiteCUAMetadata:
    dims: tuple[Platform, TaskType]  # what it holds, and where it runs
    others: dict                     # provenance: the model, the rollout's reward

Platform = "desktop" | "browser" | "mobile"
TaskType = "grounding.action" | "grounding.point" | "grounding.bbox"
           | "understanding" | "use"

class LiteUserMessage(TypedDict):    # the instruction, and the observation
    content: list[TextContent | ImageContent]

class LiteAssistantMessage(TypedDict):
    tool_calls: list[LiteToolCall]   # Lite.Gym actions, coordinates in [0, 1000]
    reasoning_content: str           # a native reasoning channel, if any
    content: list[TextContent                 # a question, an answer
                  | ActionDescriptionContent  # the action in one line
                  | InlineReasoningContent    # reasoning in the visible output
                  | HistorySummaryContent]    # a running summary of the rollout

class LiteToolMessage(TypedDict):    # what those actions returned
    tool_call_id: str
    content: list[TextContent | ImageContent]
\end{pycode}
\end{vscodebox}
\vspace{4pt}
\caption{\textbf{The \lsample{} types.}}
\label{fig:schema}
\end{figure}

\begin{figure}[h]
\begin{minipage}{\linewidth}
{\footnotesize\textbf{Grounding}: a question about a screenshot; the label is the tool call that answers it.}\par\nobreak\vspace{1pt}
\begin{vscodebox}[width=\linewidth,left=5pt,right=5pt,before skip=1pt,after skip=1pt]
\begin{jsoncode}
{"images": ["images/9c/9c4e...7b.png"],
 "metadata": {"dims": ["desktop", "grounding.point"]},
 "messages": [
   {"role": "user",
    "content": [
      {"type": "image", "index": 0},
      {"type": "text", "text": "Where is the save button?"}]},
   {"role": "assistant",
    "tool_calls": [{"id": "call_0", "type": "function",
      "function": {"name": "point", "arguments": {"coordinate": [850, 120]}}}]}]}
\end{jsoncode}
\end{vscodebox}

\vspace{11pt}
{\footnotesize\textbf{Understanding}: a question about a screenshot, answered in text, with no tool call.}\par\nobreak\vspace{1pt}
\begin{vscodebox}[width=\linewidth,left=5pt,right=5pt,before skip=1pt,after skip=1pt]
\begin{jsoncode}
{"images": ["images/1b/1b77...3d.png"],
 "metadata": {"dims": ["mobile", "understanding"]},
 "messages": [
   {"role": "user",
    "content": [
      {"type": "image", "index": 0},
      {"type": "text", "text": "What does this screen show?"}]},
   {"role": "assistant",
    "content": [{"type": "text", "text": "The music player, paused."}]}]}
\end{jsoncode}
\end{vscodebox}

\vspace{11pt}
{\footnotesize\textbf{Rollout} (\code{use}): assistant and tool turns alternate until the model answers.}\par\nobreak\vspace{1pt}
\begin{vscodebox}[width=\linewidth,left=5pt,right=5pt,before skip=1pt,after skip=1pt]
\begin{jsoncode}
{"images": ["images/3f/3fa9...c1.png", "images/b2/b20d...4e.png", "..."],
 "metadata": {"dims": ["desktop", "use"], "others": {"model_id": "gpt-5.5"}},
 "messages": [
   |\itshape\color{black!55}\# the instruction, and the screen it starts on|
   {"role": "user",
    "content": [
      {"type": "text", "text": "Set the title font to Arial."},
      {"type": "image", "index": 0}]},

   |\itshape\color{black!55}\# the model reasons, describes what it will do, and acts|
   {"role": "assistant",
    "reasoning_content": "The title is selected, so the font box applies to it.",
    "content": [{"type": "action_description", "text": "Open the font menu."}],
    "tool_calls": [{"id": "call_0", "type": "function",
      "function": {"name": "computer", "arguments": {"actions": [
        {"action": "click", "coordinate": [412, 87]}]}}}]},

   |\itshape\color{black!55}\# the screen that action produced|
   {"role": "tool", "tool_call_id": "call_0",
    "content": [{"type": "image", "index": 1}]},

   |\itshape\color{black!55}\# the next step|
   {"role": "assistant",
    "reasoning_content": "The menu is open and Arial is visible in the list.",
    "content": [{"type": "action_description", "text": "Choose Arial."}],
    "tool_calls": [{"id": "call_1", "type": "function",
      "function": {"name": "computer", "arguments": {"actions": [
        {"action": "click", "coordinate": [388, 214]}]}}}]},
   {"role": "tool", "tool_call_id": "call_1",
    "content": [{"type": "image", "index": 2}]},

   |\itshape\color{black!55}\# many more assistant and tool turns, omitted here|
   "...",

   |\itshape\color{black!55}\# the answer that ends the rollout|
   {"role": "assistant",
    "content": [{"type": "text", "text": "Done."}]}]}
\end{jsoncode}
\end{vscodebox}
\end{minipage}
\caption{\textbf{One \lsample{} per task type.}}
\label{fig:sample}
\end{figure}

\subsection{Converted Datasets}
\label{app:converted}

We convert $9$ public datasets into \lsample{}s, across desktop, browser, and mobile (Table~\ref{tab:corpora}). One converter per dataset reads the upstream release and emits \lsample{}s. Coordinates are rescaled to $[0,1000]$, actions are rebuilt in the \lgym{} action set so that key names share one vocabulary, and images are deduplicated by content. We drop, rather than repair, any \lsample{} whose image is missing or whose coordinates fall outside that range; for a \code{use} rollout we drop every step of the rollout. Each dataset's upstream release and license are recorded in the dataset card we publish.

\begin{table}[h]\centering
\caption{\textbf{The $\bm{9}$ converted datasets, in \lsample{}s by platform.} Grounding sums the \code{action}, \code{point}, and \code{bbox} sub-types.}
\label{tab:corpora}
\footnotesize
\setlength{\tabcolsep}{4pt}
\begin{tabular*}{\linewidth}{@{\extracolsep{\fill}}l l rrr r@{}}
\toprule
\textbf{Dataset} & \textbf{Platform} & \textbf{Grounding} & \textbf{Understanding} & \textbf{Rollout} & \textbf{Total} \\
\midrule
Aguvis~\citeyearpar{aguvis2024} & Desktop  &        $260$ &          $0$ &          $0$ &        $260$ \\
                      & Browser  &  $634{,}314$ &          $0$ &    $3{,}729$ &  $638{,}043$ \\
                      & Mobile   &   $57{,}498$ &          $0$ &   $17{,}820$ &   $75{,}318$ \\
ScaleCUA~\citeyearpar{scalecua2025} & Browser  &   $69{,}246$ &   $32{,}445$ &   $80{,}800$ &  $182{,}491$ \\
GUI-360~\citeyearpar{gui360_2025} & Desktop  &   $72{,}415$ &   $86{,}730$ &   $10{,}482$ &  $169{,}627$ \\
GUIOdyssey~\citeyearpar{guiodyssey2024} & Mobile   &          $0$ &  $125{,}197$ &    $8{,}330$ &  $133{,}527$ \\
GUIAct~\citeyearpar{guicourse2025} & Browser  &   $14{,}379$ &          $0$ &    $5{,}497$ &   $19{,}876$ \\
                      & Mobile   &          $0$ &          $0$ &    $7{,}438$ &    $7{,}438$ \\
OpenCUA~\citeyearpar{opencua2025} & Desktop  &          $0$ &          $0$ &   $22{,}336$ &   $22{,}336$ \\
UI-Genie-Agent~\citeyearpar{uigenie2025} & Mobile   &          $0$ &          $0$ &    $4{,}705$ &    $4{,}705$ \\
CAGUI~\citeyearpar{agentcpmgui2025} & Mobile   &          $0$ &    $2{,}715$ &        $600$ &    $3{,}315$ \\
Multimodal-Mind2Web~\citeyearpar{mind2web2023} & Browser  &          $0$ &          $0$ &        $597$ &        $597$ \\
\midrule
Total                 &          &  $848{,}112$ &  $247{,}087$ &  $162{,}334$ & $1{,}257{,}533$ \\
\bottomrule
\end{tabular*}
\end{table}

\subsection{Fresh Rollouts}
\label{app:rollouts}

\begin{table}[!h]\centering
\caption{\textbf{Rollouts with verified reward per teacher, in \lsample{}s.} --: not recorded.}
\label{tab:recorded}
\footnotesize
\setlength{\tabcolsep}{4pt}
\begin{tabular*}{\linewidth}{@{\extracolsep{\fill}}l l rrr r@{}}
\toprule
\textbf{Training environment} & \textbf{Platform} & \textbf{GPT-5.5} & \textbf{Qwen3.5-27B} & \textbf{Qwen3.8-27B} & \textbf{Total} \\
\midrule
\lscalecua{}   & Desktop          & $17{,}493$ & $17{,}539$ & $17{,}807$ & $52{,}839$ \\
\lcuagym{}     & Desktop, browser & $7{,}793$  & --         & $10{,}052$ & $17{,}845$ \\
\losworld{}    & Desktop          & $2{,}374$  & --         & $2{,}410$  & $4{,}784$ \\
WebGym         & Browser          & $3{,}143$  & $967$      & --         & $4{,}110$ \\
MobileGym      & Mobile           & $1{,}279$  & --         & --         & $1{,}279$ \\
\lcuaworld{}   & Desktop          & $728$      & --         & --         & $728$ \\
\midrule
Total          &                  & $32{,}810$ & $18{,}506$ & $30{,}269$ & $81{,}585$ \\
\bottomrule
\end{tabular*}
\end{table}

We run each model through its own harness (\S\ref{sec:harness}) on our training environments and keep the rollouts with verified reward (Table~\ref{tab:recorded}). A post-processing pass then runs over the raw logs. It strips the no-op actions \code{screenshot} and \code{wait}, and records in \code{metadata.others.exclude\_reason} why a rollout might be excluded, for example because it never finished or the agent installed software. It drops a step outright when its coordinates fall outside $[0,1000]$, its action is not in the \lgym{} action set, or its call names a tool the environment did not expose.

Every recorded step carries the model's tool calls and an \code{action\_description}. Whether it also carries reasoning, and where that reasoning comes from, differs by model:

\begin{itemize}[topsep=2pt,itemsep=1pt,parsep=0pt,leftmargin=1.2em]
\item \textbf{Qwen3.5-27B}: reasoning on, so it writes \code{reasoning\_content} natively.
\item \textbf{Qwen3.8-27B}: reasoning off, so it carries no reasoning.
\item \textbf{GPT-5.5}: its reasoning is prompted, not native. We prompt it for one \code{Thought:} line per step and store that line as \code{reasoning\_content}.
\end{itemize}

Every published rollout with reasoning therefore carries it in \code{reasoning\_content}.

\clearpage%
\section{Evaluation Details}
\label{app:eval}
\paragraph{Tasks evaluated.} We evaluate every task of ScreenSpot-Pro, WebVoyager, MiniWoB++, AndroidWorld, AndroidLab, and MobileGym, and a subset of the other six benchmarks (Table~\ref{tab:catalog} gives the full counts):
\begin{itemize}[topsep=2pt,itemsep=1pt,parsep=0pt,leftmargin=1.2em]
\item \textbf{OSWorld-G:} $510$ of $564$, without the $54$ infeasible tasks, whose target element is absent.
\item \textbf{OSWorld:} $332$ of $369$, without the $29$ infeasible tasks and the $8$ that need a Google account.
\item \textbf{OSWorld 2.1:} $99$ of $108$, without the $7$ tasks that need a simulated user, the one other task that needs a GitLab server, and the multi-phase task.
\item \textbf{WindowsAgentArena:} $138$ of the $154$ tasks with task context, without the $13$ infeasible tasks and the $3$ whose evaluator already passes on the setup state.
\item \textbf{WebArena:} $241$ of $812$, the first task of each of its $241$ intent templates.
\item \textbf{MobileWorld:} $117$ of $201$, without the $40$ tasks that need MCP tools and the $44$ that need a simulated user to answer the agent.
\end{itemize}

\paragraph{Harness settings.} Each model runs through the harness of its own release (\S\ref{sec:harness}). The harness sets how many screenshots and how much text history the model keeps, and the environment sets the screenshot resolution (Table~\ref{tab:harness}). Reasoning is off for every open-weight model except UI-Venus-2, whose official harness turns it on for every task but grounding. In \S\ref{sec:training}, Qwen3.5-4B instead keeps only the latest screenshot and runs mobile at a downsampled $720\times1600$, so its scores before SFT differ from Table~\ref{tab:main}.

\begin{table}[h]\centering
\caption{\textbf{Evaluation settings of Table~\ref{tab:main}.} Top: per model, the most recent screenshots and the text history kept in context, the same on every platform. An action summary (e.g., `Previous Actions: ') keeps earlier action descriptions in the context. When a fifth screenshot arrives, Qwen3.5 and Qwen3.8 drop the four they keep and start again from it, so they keep one to four. Bottom: per benchmark, the screenshot resolution.}
\label{tab:harness}
\footnotesize
\setlength{\tabcolsep}{6pt}
\begin{tabular}{@{}p{0.24\linewidth} >{\centering\arraybackslash}p{0.18\linewidth} >{\centering\arraybackslash}p{0.18\linewidth}@{}}
\toprule
\textbf{Model} & \textbf{Screenshots} & \textbf{Text history} \\
\midrule
\multicolumn{3}{@{}>{\columncolor{tabgroup}[0pt][0pt]}l@{}}{\textbf{Specialist CUAs}} \\
UI-TARS-1.5      & 5   & full \\
EvoCUA           & 4   & action summary \\
MAI-UI           & 3   & full \\
UI-Venus-2       & 3   & full \\
\multicolumn{3}{@{}>{\columncolor{tabgroup}[0pt][0pt]}l@{}}{\textbf{General VLMs}} \\
Qwen3-VL         & 4   & action summary \\
Qwen3.5, Qwen3.8 & 1--4 & full \\
\multicolumn{3}{@{}>{\columncolor{tabgroup}[0pt][0pt]}l@{}}{\textbf{API models}} \\
Claude Opus 5    & all & full \\
GPT models       & all & full \\
\bottomrule
\end{tabular}

\vspace{8pt}
\begin{tabular}{@{}p{0.12\linewidth} p{0.30\linewidth} >{\centering\arraybackslash}p{0.18\linewidth}@{}}
\toprule
\textbf{Platform} & \textbf{Benchmark} & \textbf{Resolution} \\
\midrule
Grounding & ScreenSpot-Pro, OSWorld-G & each image's own \\
Desktop   & OSWorld, OSWorld 2.1        & $1920\times1080$ \\
          & WindowsAgentArena         & $1280\times800$ \\
Browser   & WebVoyager, WebArena      & $1280\times720$ \\
          & MiniWoB++                 & $500\times320$ \\
Mobile    & AndroidWorld, MobileWorld & $1080\times2400$ \\
          & MobileGym                 & $1080\times2400$ \\
          & AndroidLab                & $740\times1600$ \\
\bottomrule
\end{tabular}
\end{table}

\begin{figure}[t]
\centering
\includegraphics[width=\linewidth]{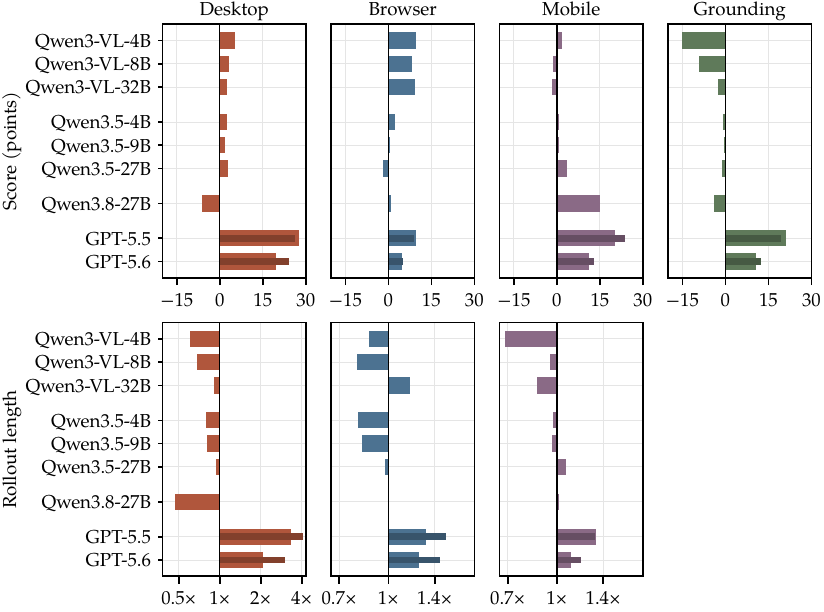}
\caption{\textbf{Effect of reasoning on score and rollout length, per model.} Figure~\ref{fig:reasoning} with Qwen3 (Qwen3-VL) and Qwen3.5 split into their sizes.}
\label{fig:reasoning-models}
\end{figure}

\paragraph{Reasoning ablation.} Figure~\ref{fig:reasoning-models} is Figure~\ref{fig:reasoning} with every model on its own row, and Table~\ref{tab:eval-full} gives the score of each model at each reasoning setting we evaluate.

\begin{table}[t]\centering
\caption{\textbf{Reference evaluation scores (\%) of the $15$ models at each reasoning setting we evaluate.} A row marked $\hookrightarrow$ reruns the model above it at another reasoning setting: reasoning on for an open-weight model, and the named reasoning effort for an API model. Qwen3-VL's reasoning rows run Qwen3-VL-Thinking, a separate checkpoint, instead of Qwen3-VL-Instruct. Qwen3.5 and Qwen3.8 switch reasoning within one checkpoint. WAA: WindowsAgentArena. --: not applicable. $\circ$: not run, because WebArena rollouts are slow. Rows marked (default), which Table~\ref{tab:main} reports, run open-weight models with reasoning off and API models at medium effort.}
\label{tab:eval-full}
\footnotesize
\setlength{\tabcolsep}{4.3pt}
\newcommand{\rot}[1]{\adjustbox{angle=45,lap=\width-1.1em}{#1}}
\begin{tabular}{@{}l cc @{\hskip 7pt} ccc @{\hskip 7pt} ccc @{\hskip 7pt} cccc@{}}
\toprule
\textbf{Model} & \rot{ScreenSpot-Pro} & \rot{OSWorld-G} & \rot{OSWorld} & \rot{OSWorld 2.1} & \rot{WAA} & \rot{WebVoyager} & \rot{WebArena} & \rot{MiniWoB++} & \rot{AndroidWorld} & \rot{AndroidLab} & \rot{MobileWorld} & \rot{MobileGym} \\
& \multicolumn{2}{c}{\scriptsize Grounding} & \multicolumn{3}{c}{\scriptsize Desktop} & \multicolumn{3}{c}{\scriptsize Browser} & \multicolumn{4}{c}{\scriptsize Mobile} \\
\cmidrule(lr){2-3}\cmidrule(lr){4-6}\cmidrule(lr){7-9}\cmidrule(lr){10-13}
\multicolumn{13}{@{}>{\columncolor{tabgroup}[0pt][0pt]}l@{}}{\textbf{Specialist CUAs}} \\
UI-TARS-1.5-7B                    & 49.8 & 60.6 & 27.5 & 0.5 & 16.7 & 46.2 & 9.1 & 18.4 & 19.8 & 36.2 & 11.1 & 13.3 \\
\addlinespace[5pt]
EvoCUA-8B                         & 43.5 & 57.3 & 35.6 & 0.7 & 16.6 & 5.8 & 10.8 & 56.0 & -- & -- & -- & -- \\
\addlinespace[5pt]
MAI-UI-8B                         & 62.2 & 70.0 & -- & -- & -- & -- & -- & -- & 61.2 & 49.3 & 29.9 & 25.8 \\
\addlinespace[5pt]
UI-Venus-2-9B                     & 71.9 & 78.2 & 50.5 & 1.4 & 41.3 & 72.9 & 30.7 & 59.2 & 62.1 & 47.8 & 59.0 & 34.4 \\
\multicolumn{13}{@{}>{\columncolor{tabgroup}[0pt][0pt]}l@{}}{\textbf{General VLMs}} \\
Qwen3-VL-4B                       & 58.3 & 65.5 & 23.6 & 0.9 & 9.5 & 20.6 & 8.7 & 47.2 & 43.1 & 36.2 & 6.8 & 13.7 \\
\quad$\hookrightarrow$ reasoning  & 40.5 & 53.1 & 29.8 & 1.2 & 18.8 & 37.2 & $\circ$ & 49.6 & 40.1 & 36.2 & 12.8 & 17.2 \\
\addlinespace[5pt]
Qwen3-VL-8B                       & 55.2 & 64.3 & 28.1 & 0.3 & 14.5 & 29.3 & 7.1 & 53.6 & 51.7 & 39.9 & 8.5 & 16.0 \\
\quad$\hookrightarrow$ reasoning  & 44.3 & 57.1 & 32.4 & 0.4 & 20.2 & 40.3 & $\circ$ & 59.2 & 46.6 & 39.1 & 7.7 & 17.2 \\
\addlinespace[5pt]
Qwen3-VL-32B                      & 57.7 & 72.0 & 34.3 & 1.2 & 24.0 & 42.8 & 14.6 & 60.8 & 62.1 & 36.2 & 17.9 & 23.4 \\
\quad$\hookrightarrow$ reasoning  & 56.4 & 68.2 & 41.5 & 0.8 & 24.6 & 50.7 & $\circ$ & 71.2 & 59.5 & 42.8 & 13.7 & 17.2 \\
\addlinespace[5pt]
Qwen3.5-4B                        & 57.7 & 62.9 & 27.0 & 0.7 & 11.6 & 12.2 & 11.3 & 28.0 & 60.3 & 34.8 & 11.1 & 17.6 \\
\quad$\hookrightarrow$ reasoning  & 57.4 & 61.8 & 29.6 & 0.8 & 16.0 & 16.6 & $\circ$ & 28.0 & 57.8 & 35.5 & 16.2 & 16.8 \\
\addlinespace[5pt]
Qwen3.5-9B                        & 58.1 & 68.2 & 38.6 & 0.4 & 13.1 & 35.0 & 15.4 & 50.4 & 62.1 & 36.2 & 10.3 & 18.0 \\
\quad$\hookrightarrow$ reasoning  & 55.3 & 70.2 & 36.6 & 0.5 & 20.9 & 36.7 & $\circ$ & 49.6 & 59.1 & 37.7 & 16.2 & 17.2 \\
\addlinespace[5pt]
Qwen3.5-27B                       & 69.5 & 74.3 & 48.9 & 2.8 & 25.4 & 51.7 & 21.9 & 68.8 & 65.1 & 40.6 & 15.4 & 26.6 \\
\quad$\hookrightarrow$ reasoning  & 65.9 & 75.5 & 48.2 & 3.2 & 34.0 & 48.1 & $\circ$ & 68.8 & 66.4 & 44.9 & 21.4 & 28.5 \\
\addlinespace[5pt]
Qwen3.8-27B                       & 75.4 & 79.6 & 59.2 & 2.4 & 37.0 & 55.4 & 24.5 & 64.8 & 64.7 & 51.4 & 32.5 & 40.2 \\
\quad$\hookrightarrow$ reasoning (xhigh)& 73.1 & 74.1 & 44.5 & 0.5 & 34.8 & 62.2 & $\circ$ & 60.0 & 73.3 & 54.3 & 63.2 & 58.6 \\
\multicolumn{13}{@{}>{\columncolor{tabgroup}[0pt][0pt]}l@{}}{\textbf{API models}} \\
Claude Opus 5                     & 85.3 & 89.2 & 78.7 & 63.1 & 72.4 & 90.4 & 51.7 & 91.2 & 81.0 & 62.3 & 78.6 & 70.2 \\
\addlinespace[5pt]
GPT-5.5 (none)                    & 45.2 & 77.3 & 48.1 & 5.8 & 39.2 & 70.9 & $\circ$ & 72.8 & 62.5 & 52.9 & 42.7 & 34.8 \\
\quad$\hookrightarrow$ medium (default)& 74.5 & 90.4 & 71.1 & 38.6 & 66.6 & 78.4 & 47.1 & 84.8 & 78.4 & 60.9 & 80.3 & 53.9 \\
\quad$\hookrightarrow$ xhigh      & 71.1 & 90.2 & 74.8 & 33.2 & 64.3 & 76.5 & $\circ$ & 84.8 & 77.6 & 63.0 & 84.6 & 62.5 \\
\addlinespace[5pt]
GPT-5.6 Sol (none)                & 72.4 & 79.8 & 60.3 & 17.0 & 50.7 & 75.7 & $\circ$ & 76.8 & 78.4 & 56.5 & 60.7 & 46.1 \\
\quad$\hookrightarrow$ medium (default)& 82.5 & 91.2 & 73.3 & 43.9 & 70.0 & 79.8 & 53.3 & 82.4 & 75.9 & 60.9 & 86.3 & 64.1 \\
\quad$\hookrightarrow$ xhigh      & 84.2 & 92.7 & 74.8 & 54.0 & 71.6 & 79.2 & $\circ$ & 83.2 & 78.4 & 63.8 & 82.9 & 68.8 \\
\addlinespace[5pt]
GPT-6 Astra                       & 93.9 & 92.5 & 75.1 & 63.0 & 63.4 & 84.3 & 53.3 & 87.2 & 84.5 & 60.9 & 85.5 & 80.9 \\
\bottomrule
\end{tabular}
\end{table}

\clearpage%
\section{Training Details}
\label{app:training}
\paragraph{SFT.} Every SFT run in \S\ref{sec:sft} uses the same hyperparameters on every platform: two epochs over the training set, a global batch of $32$ rollouts, and the Adam optimizer ($\beta_1=0.9$, $\beta_2=0.95$, weight decay $0.1$) with a learning rate of $5\times10^{-6}$, linear warmup over the first $10\%$ of steps, and cosine decay to $10^{-6}$. Unless an ablation varies them, runs keep one screenshot ($1920\times1080$ on desktop, $1280\times720$ on browser, $720\times1600$ on mobile) and the full text history. The loss is averaged over the model's output tokens. We evaluate the checkpoint after the second epoch three times at temperature $0$ on the held-out tasks, and report the mean evaluation score.

\begin{table}[!h]\centering
\caption{\textbf{Evaluation scores (\%) of Qwen3.5-4B in every setting of \S\ref{sec:sft}.} The table extends Table~\ref{tab:sft} with the data of Figure~\ref{fig:sft-ablation}. Screenshots: the number kept in context. --: not run.}
\label{tab:sft-full}
\footnotesize
\setlength{\tabcolsep}{6pt}
\begin{tabular}{@{}l c cc c c@{}}
\toprule
 & & \multicolumn{2}{c}{\textbf{Desktop}} & \textbf{Browser} & \textbf{Mobile} \\
 & & \multicolumn{2}{c}{\losworld{}} & WebVoyager & MobileGym \\
\cmidrule(lr){3-4}
 & \textbf{Screenshots} & $1920\times1080$ & $1280\times720$ & $1280\times720$ & $720\times1600$ \\
\midrule
\multicolumn{6}{@{}>{\columncolor{tabgroup}[0pt][0pt]}l@{}}{\textbf{Without SFT}} \\
w/o reasoning & 1 & 22.6  & 14.9 & 4.4  & 14.6 \\
                   & 4 & --    & 25.0 & 8.3  & 15.2 \\
w/\phantom{o} reasoning & 1 & 27.1  & 22.6 & 14.8 & 13.9 \\
                        & 4 & --    & 25.9 & 13.5 & 16.1 \\
\multicolumn{6}{@{}>{\columncolor{tabgroup}[0pt][0pt]}l@{}}{\textbf{SFT on GPT-5.5 rollouts}} \\
w/o reasoning           & 1 & 37.2 & --   & 48.7 & 21.4 \\
w/\phantom{o} reasoning & 1 & 44.8 & 36.9 & 60.0 & 26.8 \\
                             & 4 & --   & 40.2 & 59.6 & 27.2 \\
\multicolumn{6}{@{}>{\columncolor{tabgroup}[0pt][0pt]}l@{}}{\textbf{SFT on Qwen3.5-27B rollouts}} \\
w/o reasoning           & 1 & 34.5 & -- & -- & -- \\
w/\phantom{o} reasoning & 1 & 35.4 & -- & -- & -- \\
\bottomrule
\end{tabular}
\end{table}

\paragraph{GRPO.} Every GRPO run in \S\ref{sec:rl} starts from the checkpoint SFTed on GPT-5.5 rollouts with reasoning and one screenshot, and uses the same hyperparameters on every platform: $16$ training tasks per GRPO step, $8$ rollouts per task at temperature $1$, $8$ gradient updates per GRPO step, clip ratios of $0.2$ and $0.28$, no KL penalty, and the Adam optimizer ($\beta_1=0.9$, $\beta_2=0.98$, weight decay $0.1$) with a constant learning rate of $10^{-6}$. The platforms differ in training length, reward, and evaluation:
\begin{itemize}[topsep=2pt,itemsep=1pt,parsep=0pt,leftmargin=1.2em]
\item \textbf{Desktop}: $20$ GRPO steps with $24$ concurrent environments. The reward is the \losworld{} verifier's score, which gives partial credit on some tasks. Every two GRPO steps, we evaluate each held-out task four times at temperature $1$.
\item \textbf{Browser}: $30$ GRPO steps with $24$ concurrent environments. The reward is $1$ if WebVoyager's GPT-4.1 judge accepts the rollout and $0$ otherwise. Every five GRPO steps, we evaluate each held-out task once at temperature $0$.
\item \textbf{Mobile}: $30$ GRPO steps with $32$ concurrent environments. The reward is $1$ if the task succeeds, and otherwise half the fraction of its goal checks that pass. Every five GRPO steps, we evaluate each held-out task four times at temperature $1$. A task family is the tasks of one app whose names share a prefix, at levels L1--L3; we drop L4, the hardest level, which takes $39\%$ of the step budget but succeeds on only $4\%$ of its tasks.
\end{itemize}

\end{document}